\documentclass[12pt, number, 1p]{elsarticle}
\usepackage{lineno}
\modulolinenumbers[1]
\usepackage[utf8]{inputenc} %
\usepackage[T1]{fontenc}    %
\usepackage{hyperref}       %
\usepackage{url}            %
\usepackage{booktabs}       %
\usepackage[normalem]{ulem}
\useunder{\uline}{\ul}{}
\usepackage{amsfonts}       %
\usepackage{amsmath}
\usepackage{nicefrac}       %
\usepackage{microtype}      %
\usepackage{lipsum}		%
\usepackage{graphicx}
\usepackage[numbers]{natbib}
\usepackage{doi}
\usepackage{xcolor}
\usepackage{marginnote}
\usepackage{multirow}
\usepackage{tablefootnote}
\usepackage[justification=centering]{caption}
\usepackage{subcaption}
\usepackage[backgroundcolor=black!20,linecolor=black!20,textsize=tiny]{todonotes}
\usepackage[section]{placeins}
\makeatletter
\AtBeginDocument{%
  \expandafter\renewcommand\expandafter\subsection\expandafter{%
    \expandafter\@fb@secFB\subsection
  }%
}
\makeatother

\newcommand{\todored}[1]{\todo[color=red!40]}

\begin{document}
    \begin{frontmatter}
    
    \title{Video-based Automatic Lameness Detection of Dairy Cows using Pose Estimation and Multiple Locomotion Traits}

    \author[1]{Helena Russello
	\corref{correspondingauthor1}
    }
	\cortext[correspondingauthor1]{Corresponding authors}
	\ead{helena.russello@wur.nl}
    
    \author[1]{Rik van der Tol}
    \author[2]{Menno Holzhauer}
    \author[1]{Eldert J. van Henten}
    \author[1]{Gert Kootstra
	\corref{correspondingauthor1}
    }
    \ead{gert.kootstra@wur.nl}
 	
    \address[1]{Agricultural Biosystems Engineering group, Wageningen University \& Research, Wageningen, The Netherlands }
    \address[2]{Ruminant Health Department, Royal GD AH, Deventer, The Netherlands}

    \begin{abstract}
This study presents an automated lameness detection system that uses deep-learning image processing techniques to extract multiple locomotion traits associated with lameness.
Using the T-LEAP pose estimation model, the motion of nine keypoints was extracted from videos of walking cows.
The videos were recorded outdoors, with varying illumination conditions, and T-LEAP extracted 99.6\% of correct keypoints. 
The trajectories of the keypoints were then used to compute six locomotion traits: back posture measurement, head bobbing, tracking distance, stride length, stance duration, and swing duration. 
The three most important traits were back posture measurement, head bobbing, and tracking distance.
For the ground truth, we showed that a thoughtful merging of the scores of the observers could improve intra-observer reliability and agreement.
We showed that including multiple locomotion traits improves the classification accuracy from 76.6\% with only one trait to 79.9\% with the three most important traits and to 80.1\% with all six locomotion traits. ~\\
\end{abstract}

    	
    \end{frontmatter}

%\linenumbers 
\section{Introduction}

Lameness is a painful gait disorder in dairy cows and is often characterized by abnormal locomotion of the cow. 

A recent literature review~\citep{thomsen2023prevalence} estimated the global prevalence of lameness at 22.8\%, with little change in the last 30 years.
Lameness has a negative impact on welfare~\citep{whay2017impact} and leads to substantial economic losses~\citep{enting1997economic} due to decreased milk production and reproduction~\citep{huxley2013impact} as well as premature culling~\citep{enting1997economic}.
While lameness is commonly assessed by trained observers performing visual locomotion scoring of the herd, the procedure is time-consuming and cannot realistically be performed on a regular basis. 
Hence, dairy farms could benefit from automatic lameness detection.

To date, a number of studies have investigated ways to automate locomotion scoring and lameness detection using camera systems. Video cameras are an attractive sensor for this application as they are relatively inexpensive, non-intrusive, and scale well with large herds. 
A three-step approach is commonly taken to detect lameness from videos: (1) use computer vision methods to localize body parts of interest, (2) compute one or more locomotion traits from the extracted body parts, and (3) train a classifier to score lameness using the locomotion traits as features.
In the past, the body parts were localized using classical computer vision methods such as background subtraction~\citep{song_automatic_2008, poursaberi_real-time_2010, zheng_cows_2023, zhao_automatic_2023}. These methods worked in experimental settings but were sensitive to changes in background and light, making them less applicable in practice.
Others placed physical markers (tags or paint marks) on the cows' body parts and tracked the markers with specialized software~\citep{blackie_associations_2013, karoui_deep_2021}. In practical settings, however, physical markers don't scale well to large herds as they need to be placed on each cow and cleaned regularly to remain visible.
More recently, with the emergence of deep neural networks, studies started using deep-learning-based object detection~\citep{wu_lameness_2020, kang_accurate_2020, jiang_dairy_2022, zheng_cows_2023} to localize the legs or the back of the cows, object segmentation~\citep{arazo_segmentation_2022} to extract the body contour from the background, or markerless (i.e., without physical markers) pose estimation~\citep{mathis_deeplabcut_2018, russello_t-leap_2021, zhao_automatic_2023, barney2023deep, taghavi2023cow} to localize multiple body parts in videos. Although they typically require more data than classical approaches, the deep-learning methods cope well with complex background and light conditions and can sometimes even cope with occlusions such as fences~\citep{russello_t-leap_2021, taghavi2023cow}.

Once localized in the images or video frames, the outline of the spine, for instance, can be used to compute the back posture~\citep{poursaberi_real-time_2010, viazzi_analysis_2012, van_hertem_automatic_2014, viazzi_comparison_2014, van_hertem_implementation_2018, jiang_dairy_2022, zheng_cows_2023}, and the location of the legs to compute the tracking distance~\citep{song_automatic_2008, blackie_associations_2013} or stride length~\citep{blackie_associations_2013, wu_lameness_2020, zheng_cows_2023}.
To the best of our knowledge, almost all studies on lameness detection from videos use only one locomotion trait as a feature to score lameness, and so far, only~\citep{zhao2018automatic}, \citep{barney2023deep}, and \citep{zhao_automatic_2023} combined multiple locomotion traits.

Using the locomotion trait(s) as feature(s), supervised learning classifiers can then be trained to score lameness.
In supervised learning, classifiers learn from given examples, also known as ground truth or golden standard. Manual locomotion scores, that is, locomotion scores provided by one or more observers, make up the ground truth of lameness detection classifiers.
The subjective nature of manual locomotion scoring is a well-known problem~\citep{schlageter-tello_effect_2014} and often leads to low intra- and inter-observer reliability and agreement. 
However, a classifier can only be as good as its ground truth, so information about the reliability of the locomotion scale is necessary. 
Yet, observer reliability and agreement are seldom reported, let alone analyzed.

Three critical gaps emerge from the studies discussed so far: 
(1) the use of obsolete image processing methods remains frequent, 
(2)few studies combine multiple locomotion traits for lameness classification,
and (3) the reliability of the ground truth is seldom reported. 
This paper addresses the three gaps mentioned above and proposes a non-intrusive and fully automated approach to camera-based lameness detection that includes multiple locomotion traits.
Additionally, the code of this paper is open-source\footnote{Code available at: \url{https://github.com/hrussel/lameness-detection}}.

We used videos of walking cows that were scored on a 5-point locomotion scoring scale by four observers.
We first reported and discussed the intra- and inter-observer reliability and agreement of the ground truth. 
We merged scores from the multiple observers to a binary scale.
We then trained T-LEAP~\citep{russello_t-leap_2021}, a deep-learning markerless pose estimation model, to automatically extract the motion of multiple body parts (later referred to as keypoints) from videos of walking cows. 
The sequences of keypoints were used to compute six locomotion traits that are known to be correlated with locomotion scores~\citep{schlageter-tello_relation_2015}, namely back posture measurement, head bobbing, tracking distance, stride length, stance duration, and swing duration.
Using the locomotion traits mentioned above as input features, we trained multiple machine-learning models to classify the gait as \textit{normal} or \textit{lame}.
We evaluated the performance of each model and showed the impact of using different combinations of locomotion traits on the lameness classification.

\section{Materials}

\subsection{Data acquisition}\label{subsec:data-acq}

The data were collected in Tilburg, The Netherlands, at a commercial dairy farm whose herd contained about a hundred Holstein-Frisian cows.
The data were collected between 9 am and 4 pm on 8 different days between May and July 2019. %
The cows were filmed from the side while they walked freely through an outdoor passageway.
A ZED RGB-D stereo camera\footnote{\url{https://www.stereolabs.com/zed-2/}} was placed 2 meters above the ground, at 4.5m from the fence of the passageway. 
The camera directly faced the passageway and recorded in landscape mode at Full-HD (1080p) resolution at 30 frames per second.
The recordings were saved into short videos of about 7.6 seconds, which was the average time a cow needed to walk the visible part of the passageway (9.5 meters).
The same data acquisition campaign was used by~\cite{russello_t-leap_2021} on the same farm.  
In total, 1101 videos were collected, and a subset of 272 videos were selected according to the following criteria: there was only one cow on the passageway, and the cow walked from the left to the right without distraction or interruption.

During the data collection, no process was set in place to automatically link the videos to an individual cow (e.g., by means of an RFID tag reader). The cows were, therefore, assigned a unique identifier at a later time by manually grouping the individual cows. We identified 98 unique cows, out of which 24 cows were present in the videos only once, 21 cows twice, 25 three times, 17 four times, 6 five times, 3 six times, 1 seven times, and 1 eight times. For the cows that were present multiple times, some were recorded at different times on the same day, and some on different days.

\subsection{Locomotion scoring}

The locomotion scoring was performed using the 5-point discrete scale described by Sprecher et al. 1997~\citep{sprecher_lameness_1997}, where a score of 1 corresponds to \textit{normal gait}, 2 to \textit{midly lame}, 3 to \textit{moderately lame}, 4 to \textit{lame} and 5 to \textit{severely lame}.
Four observers scored the videos: one expert (A) with 20 years of experience in visual locomotion scoring and three observers (B, C, D) with no prior experience in locomotion scoring but with a background in animal science and dairy farming. 
The inexperienced observers were trained by the expert (A) before the scoring session. 
During the scoring session, each video was played twice in a row to give enough time to observe the locomotion. 
To ensure consistency, the observers were asked to give the lowest score if they were hesitating between two scores.
All the videos were scored on the same day. 
After the scoring session, the observers indicated no cow recognition, i.e., that they did not recognize the individual cows that appeared in multiple videos. 
Table~\ref{tab:scores-distrib} shows the distribution of the scores assigned by the four observers. 
The distribution of the scores was highly imbalanced and indicated a homogeneous herd, where most cows were distributed throughout the first two levels of the scale (normal, mildly lame), which is typical of herds with a low prevalence of lameness~\citep{thomsen2008evaluation}.

\begin{table}[htbp]
    \centering
    
    \caption{Distribution of the locomotion scores assigned by the observers}
    \label{tab:scores-distrib}
    \begin{tabular}{c|ccccc|c}
        \hline
        \multirow{2}{*}{Observer} & \multicolumn{5}{c|}{Locomotion score} & \multirow{2}{*}{Total}\\
                  & 1    & 2    & 3    & 4    & 5    \\
        \hline
        A & 115 & 99 & 27 & 31 & 0 & 272 \\
        B & 109 & 80 & 54 & 26 & 3 & 272 \\
        C & 101 & 119 & 34 & 15 & 3 & 272 \\
        D & 141 & 80 & 38 & 12 & 1 & 272\\
        \hline
        Distribution & 42.8\% & 34.7\% & 14.1\% & 7.7\% & 0.6\% & \\
        \hline
    \end{tabular}
    
\end{table}

\subsection{Observers reliability and agreement}

Manual locomotion scoring is subjective~\citep{flower_effect_2006}. Investigating the reliability and agreement between (inter-rater) and among (intra-rater) raters can inform on the quality of the data.
Reliability estimates the capability of the raters to differentiate between the different scores, whereas agreement assesses the capability of the raters to assign the same score to the same data point. 
Reliability was measured with Krippendorff's $\alpha$~\citep{krippendorff2011computing} for ordinal values, and agreement was presented as the Percentage of Agreement (PA). 
The inter-observer and intra-observer measures are reported in Table~\ref{tab:inter-intra-obs}.
Note that, as the videos were only scored once, the intra-observer agreement was estimated by comparing the scores between pairs of videos of the same cows recorded at less than 48-hour intervals. 

\begin{table}[]
   
    \centering
    \caption{Inter-observer (A,B,C,D) and intra-observer reliability and agreement of the locomotion scoring.}
    \label{tab:inter-intra-obs}
    \begin{tabular}{l|c|cccc}
    \hline
    Observer & (A,B,C,D)        & A       & B      & C      & D      \\ \hline
    Krippendorff's $\alpha$        &      0.602          &    0.611     &  0.552      &  0.653      &  0.585      \\
    Percentage of Agreement       &      55.8          &   56.4      &   49.1     &    60.0    &     58.2   \\ \hline
    \end{tabular}
    
\end{table}

\subsection{Merging the locomotion scores}

In order to create the ground truth for the classification dataset, each sample (video) was assigned a single ground-truth label based on the locomotion scores provided by multiple observers. 
To achieve this, the scores from the four observers were combined into one value by calculating the rounded-down average of the two observers with the highest intra-observer reliability.

The majority of the studies on lameness detection focus on 2-level (normal, lame) or 3-level (normal, moderately lame, lame) locomotion scales rather than on a 5-level scale \citep{sprecher_lameness_1997, flower_effect_2006}. 
The main motivation for resorting to smaller resolutions in locomotion scales is the distribution of the scores throughout the scale. 
Severely lame cows are rare, as most get treatment or are culled before they reach this level of lameness~\citep{engel_assessment_2003}. 
This results in a heavily unbalanced score distribution, most scores being levels 1, 2, and 3. 
It is then challenging to train a classifier on unbalanced datasets, especially when little examples are available for some classes. 
As shown in Table~\ref{tab:scores-distrib}, the distribution of the locomotion scores was highly imbalanced.
Therefore, in order to balance the dataset, we merged the levels of the scale into a binary scale where level 1 indicated a \textit{normal} gait, and levels 2,3,4 and 5 indicated a \textit{lame} gait.
This resulted in an agreement of 80\%, and reliability of 0.590.
Note that reliability metrics such as Krippendorff's $\alpha$ can decrease when the scoring scale is smaller because the chance of agreement is larger.

After combining the scores of the multiple observers and turning to a binary scale, the ground truth for the classification dataset, consisting of 272 videos, contained 143 videos labeled as \textit{normal}, and 129 videos labeled as \textit{lame}.

\section{Methods}

\begin{figure}[ht]
    \centering
    
    \includegraphics[width=0.8\linewidth]{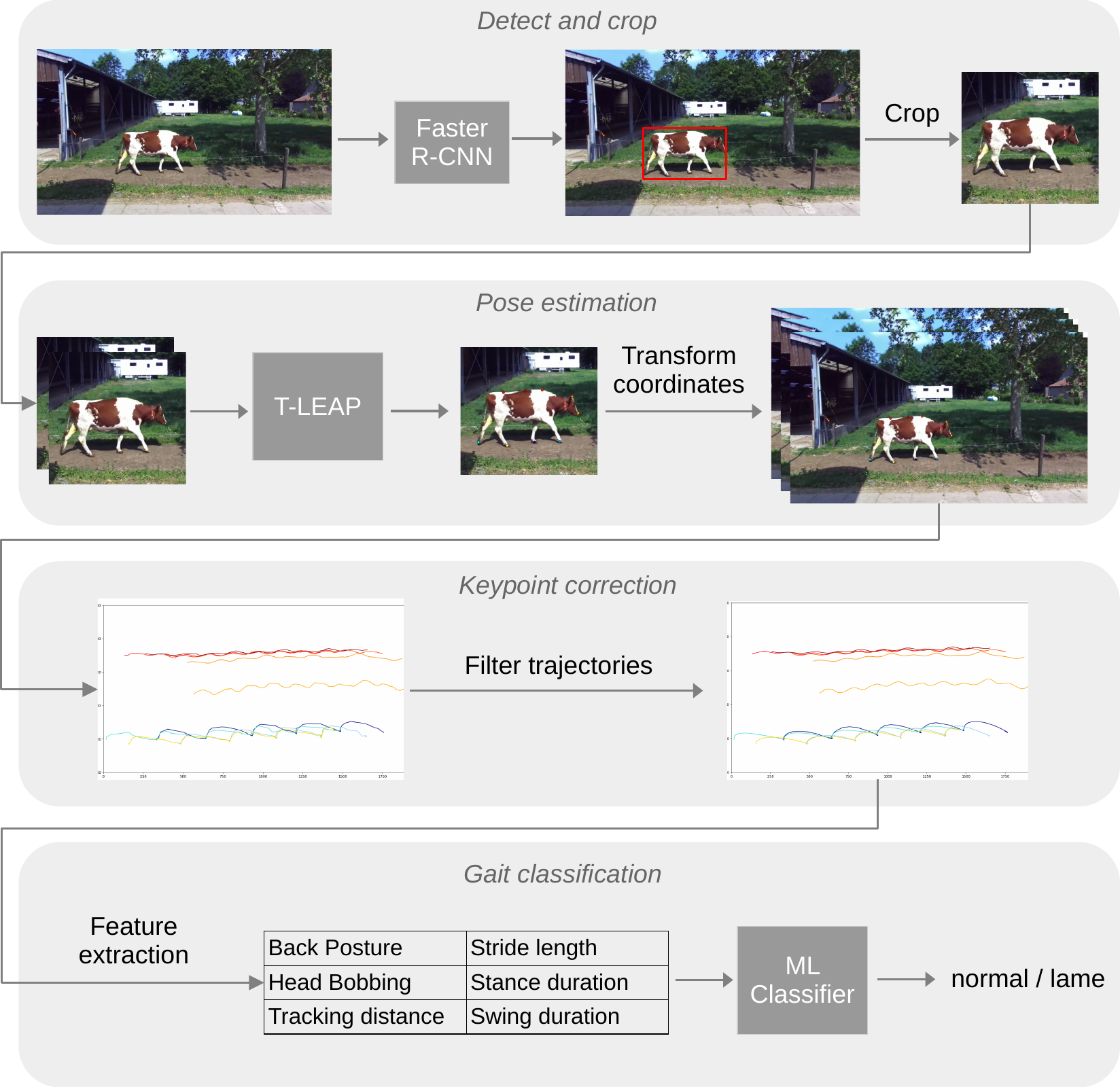}
    \caption{Summary of the video processing procedure.}
    \label{fig:pose-summary}
\end{figure}

Our methodology consisted of three main parts: pose estimation, gait features extraction, and lameness classification. These parts are described in detail in the following subsections, and a graphical summary of the methods is provided in Figure~\ref{fig:pose-summary}.

\subsection{Pose estimation}

Pose estimation models can be used to predict the position of keypoints (body parts) in images and videos without requiring physical markers. 
T-LEAP is a recent, deep-learning-based, temporal pose estimation model that was trained to detect keypoints on the body of cows in videos~\citep{russello_t-leap_2021}. 
The model used sequences of successive frames to predict the coordinate of the keypoints, and was shown to perform better than static approaches in the presence of occlusions (such as fences).
In this study, we used T-LEAP to extract nine keypoint coordinates from the video frames (Figure~\ref{fig:9keypoints}). 
In the next paragraphs, we describe the steps necessary for image cropping, pose estimation, and correction.

\subsubsection{Detect-and-crop}\label{subsec:detect-crop}
The T-LEAP model required the input frames to be square and cropped around the cow's body. 
The cows were automatically localized in the video frames using the Faster Region-based Convolutional Neural Network (Faster R-CNN), an object-detection model that returns the coordinates of a bounding box (bbox) around each object of interest (here, cows).
We used the Faster R-CNN model (with ResNeXt-101 backbone) trained on the COCO-2017 dataset from the \textit{Detectron2} library~\citep{wu2019detectron2}.
The COCO-2017 dataset contained 118K training images with annotations for 80 categories of objects, among which 8014 bounding-box annotations of cows.
The Faster R-CNN model from \textit{Detectron2} worked out of the box and could detect the cows in our video frames without fine-tuning.
Each frame of each video was fed to the object-detection model, which returned a list of bounding boxes, one for each detected cow. 
For each frame, the bounding box was made square by extending the top and bottom coordinates to match the width while keeping the cow vertically centered. 
A 100-pixel padding was added to all four sides to ensure that the body of the cow was fully visible in the cropped area.
The image was cropped to the coordinates of the extended bounding box and re-scaled to a size of $200\times200$ pixels.
The coordinates of the cropping bounding box were saved to transform the keypoint predictions back to the true coordinates for the video frame.

\subsubsection{Keypoint detection}
\begin{figure}[ht]
    \centering
    \includegraphics[]{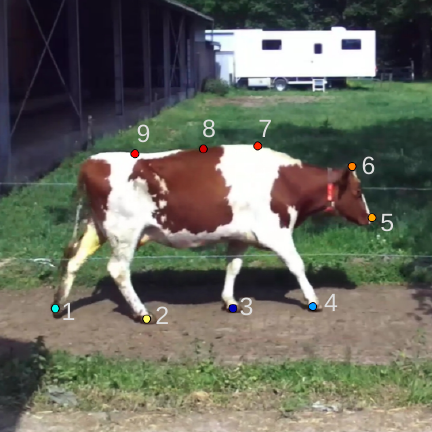}
    \caption{The 9 keypoints (anatomical landmarks) as described in~\cite{russello_t-leap_2021}. The keypoints are named as follows: 1: Left-hind hoof, 2: Right-hind hoof, 3: Left-front hoof, 4: Right-front hoof 5: Nose, 6: Forehead, 7: Withers, 8: Caudal thoracic vertebrae, 9: Sacrum.}
    \label{fig:9keypoints}
\end{figure}

We trained T-LEAP to predict the location of 9 keypoints. They represented the location of the following anatomical landmarks: Nose, Forehead, Withers, Sacrum, Caudal thoracic vertebrae, and the four Hooves (Figure~\ref{fig:9keypoints}). 
The location of these nine keypoints was needed for extracting the gait features described in subsection~\ref{subsec:feat-extr}.
T-LEAP was trained with sequences of 2 consecutive frames as input because the authors reported the best performance with T=2~\citep{russello_t-leap_2021}.

A pose estimation dataset was created for training and evaluating T-LEAP, using 28 videos of unique cows randomly selected from of the 272 available videos. 
The coordinates of the nine keypoints were annotated for each frame of the 28 videos and divided into 968 non-overlapping sequences of 2 frames. We refer to each set of consecutive frames as a sample.
T-LEAP was trained with a random subset of 80\% of the samples (i.e., 774 training samples) and evaluated on the remaining 20\% of the samples (i.e., 194 test samples).
We used the same training procedure and hyper-parameters settings as described in the original T-LEAP paper~\citep{russello_t-leap_2021}.

The trained T-LEAP model was then used to predict the location of the nine keypoints on all 272 videos of walking cows, including the 28 videos used for training.
Each video frame was cropped around the body of the cow, and sequences of 2 consecutive frames were fed to the pose estimation model. 
The keypoint coordinates predicted by the model were then transformed to the true coordinates of the video. 
For each video, this resulted in the coordinates $(x_t,y_t)$ of each keypoint for each frame $t$. 
We refer to the collection of keypoints of one video as "keypoints trajectories". In essence, these trajectories represent the motion of the anatomical landmarks localized by the pose-estimator in the 2D image plane.

\subsubsection{Keypoint correction}

\begin{figure}[ht]
    \centering
    \begin{subfigure}[b]{0.5\textwidth}
        \centering
        \includegraphics[width=1\textwidth]{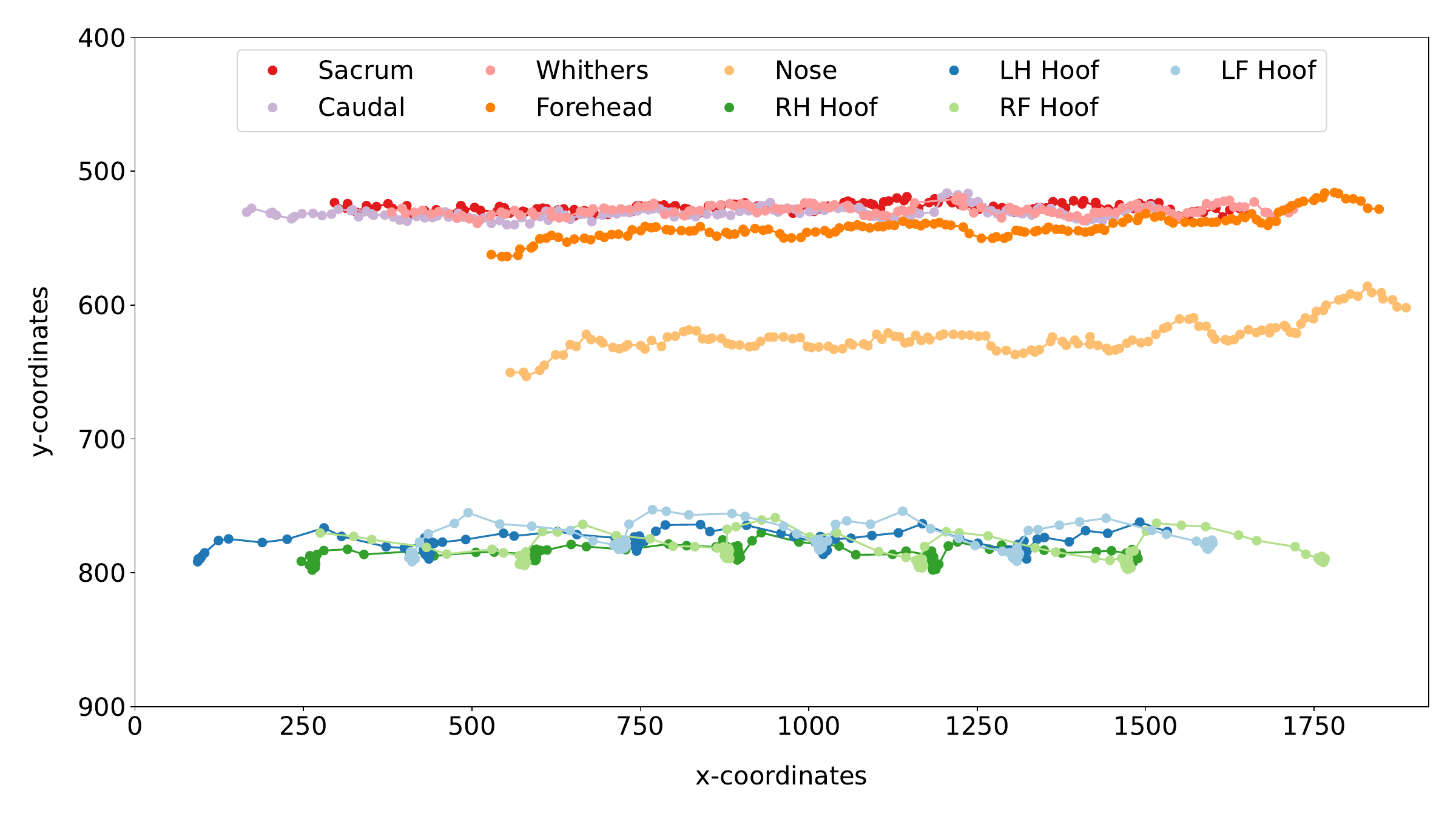}
        \caption{Normal gait, unfiltered}
    \end{subfigure}%
    \begin{subfigure}[b]{0.5\textwidth}
        \centering
        \includegraphics[width=1\textwidth]{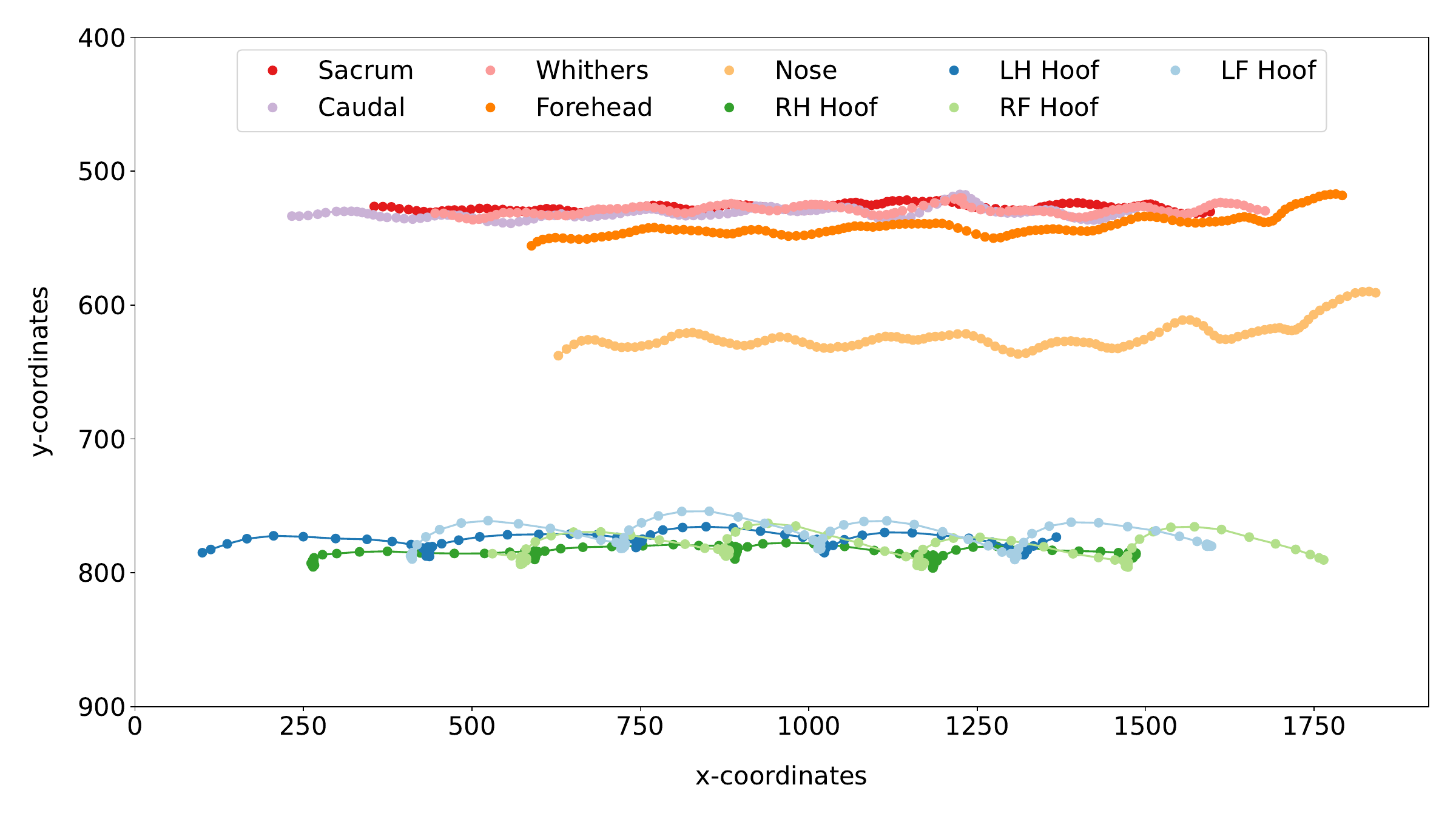}
        \caption{Normal gait, filtered}
    \end{subfigure}
    \begin{subfigure}[b]{0.5\textwidth}
        \centering
        \includegraphics[width=1\textwidth]{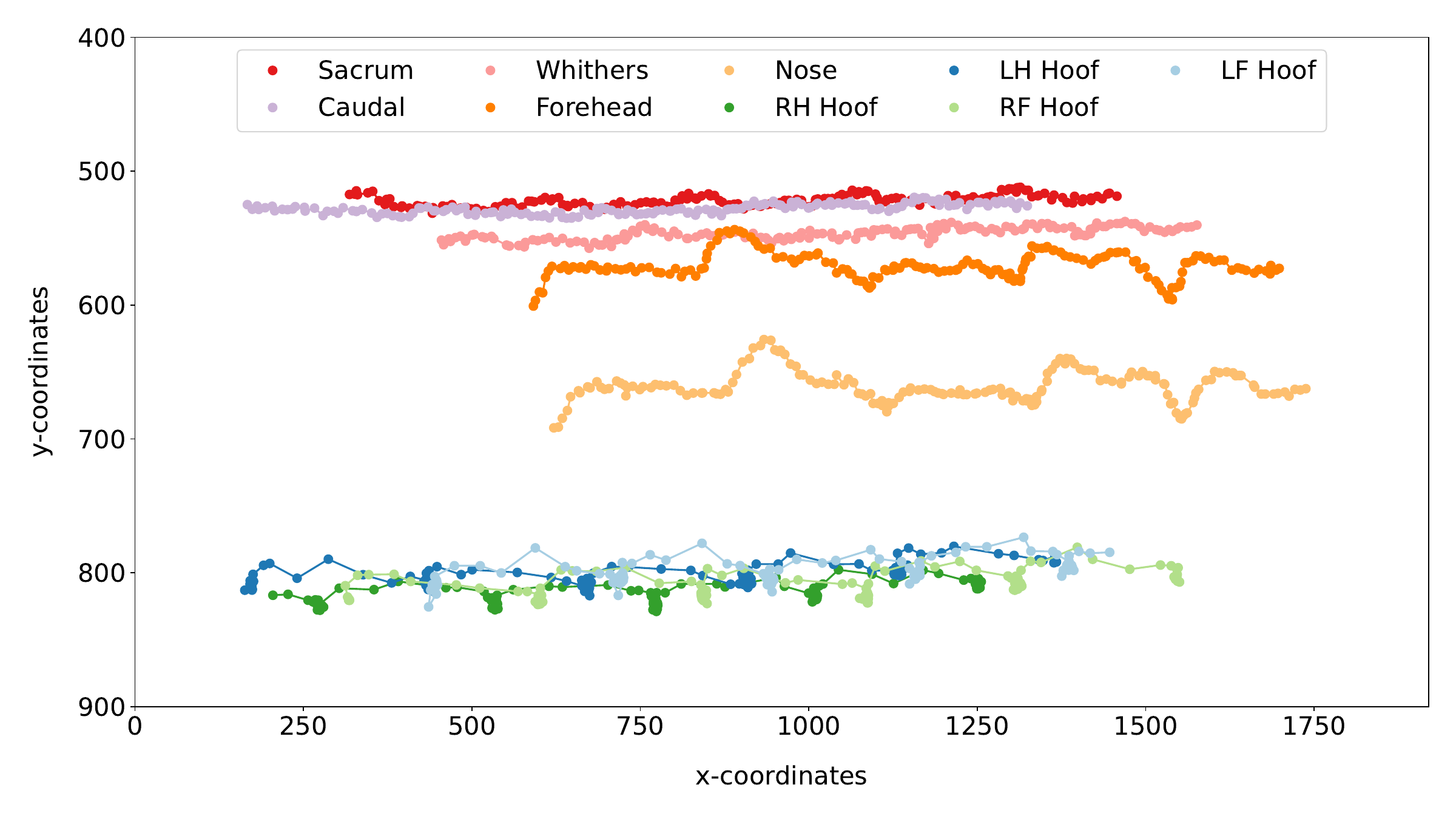}
        \caption{Lame gait, unfiltered}
    \end{subfigure}%
    \hfill
    \begin{subfigure}[b]{0.5\textwidth}
        \centering
        \includegraphics[width=1\textwidth]{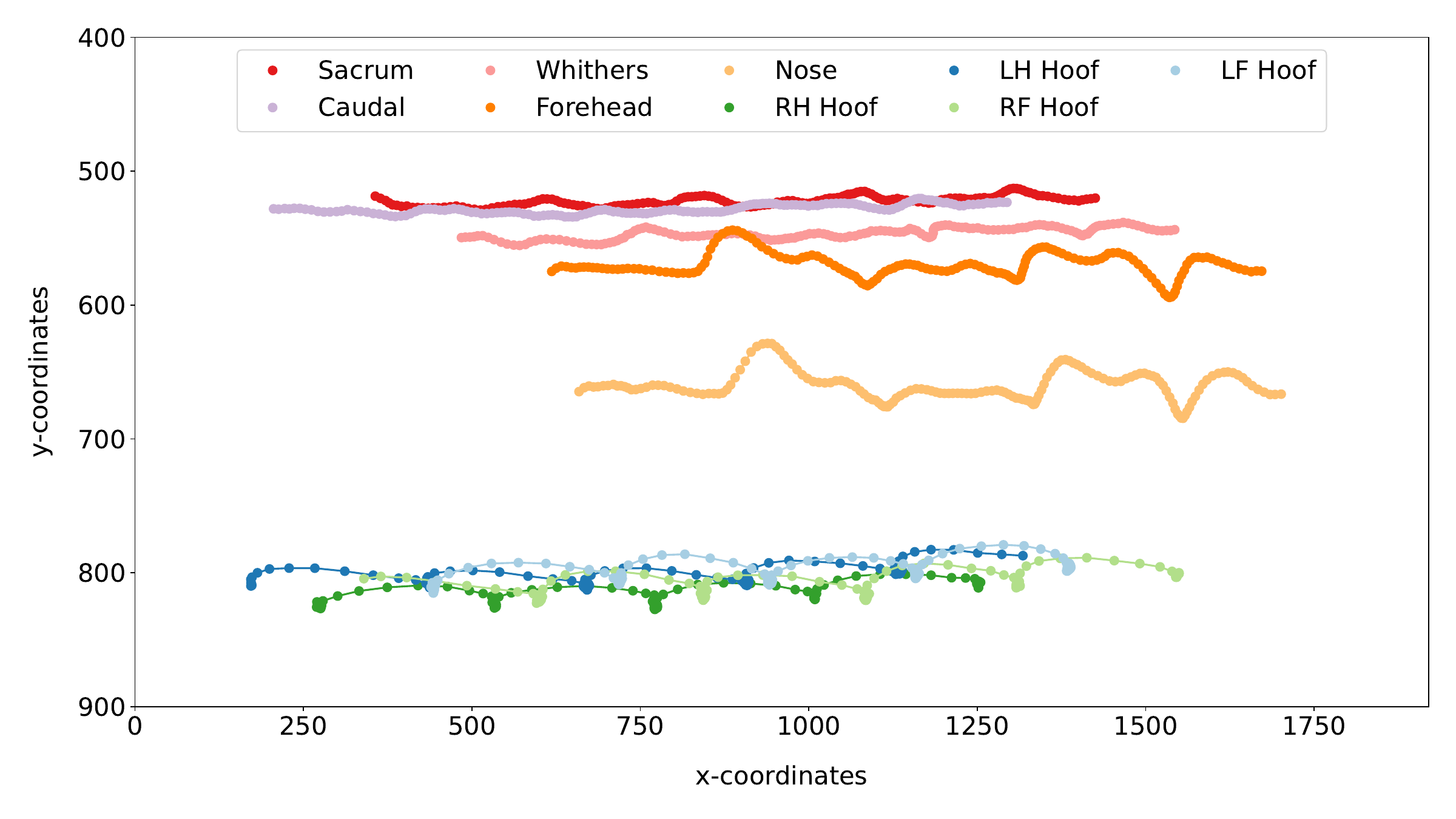}
        \caption{Lame gait, filtered}
    \end{subfigure}%
    
    \caption{Example of the keypoint trajectories extracted with T-LEAP (left), and after filtering (right) for a normal gait (top) and a lame gait (bottom).}
    \label{fig:trajectory-example}
\end{figure}

In our set of 272 videos, we identified 98 individual cows.
There were 28 videos of unique cows included in training the pose estimation model, and thus 70 cows that the pose estimation model did not see.
In their generalization experiment, the authors of T-LEAP reported a percentage of correct keypoints (PCKh@0.2) of $93.8\%$ on known cows (i.e., cows included in the training set) and a performance of $87.6\%$ on unknown cows (i.e., cows not included in the training set). 
It was, therefore, expected to have errors in the predicted keypoint trajectories. 
To deal with that, we developed a method for correcting the keypoints. 
First, to identify and correct large outliers in the trajectories, we used a Median-Absolute-Deviation (MAD) filter with a temporal window of size 3. We then applied a Savitzky–Golay filter~\citep{savitzky1964smoothing} (window=10, order=3) to smooth the trajectories temporally.
Figure~\ref{fig:trajectory-example} shows examples of trajectories with outliers before and after applying the filters.

\subsection{Gait features extraction}\label{subsec:feat-extr}

Using the keypoint trajectories, we computed six locomotion traits that were shown to be correlated with locomotion scores~\citep{schlageter-tello_relation_2015}, namely Back Posture Measurement (BPM), Head Bobbing Amplitude (HBA), Tracking distance (TRK), Stride Length (STL), Stance Duration (STD) and Swing Duration (SWD). 
All features relied on step detection, that is, knowing when each hoof was moving (swing phase) or remained still (stance phase). 
Hence, in the following paragraphs, we first describe the implementation of the step detection, followed by the implementation of the gait features.

\subsubsection{Step detection}
\begin{figure}[ht]
    \centering
    \includegraphics[width=0.8\textwidth]{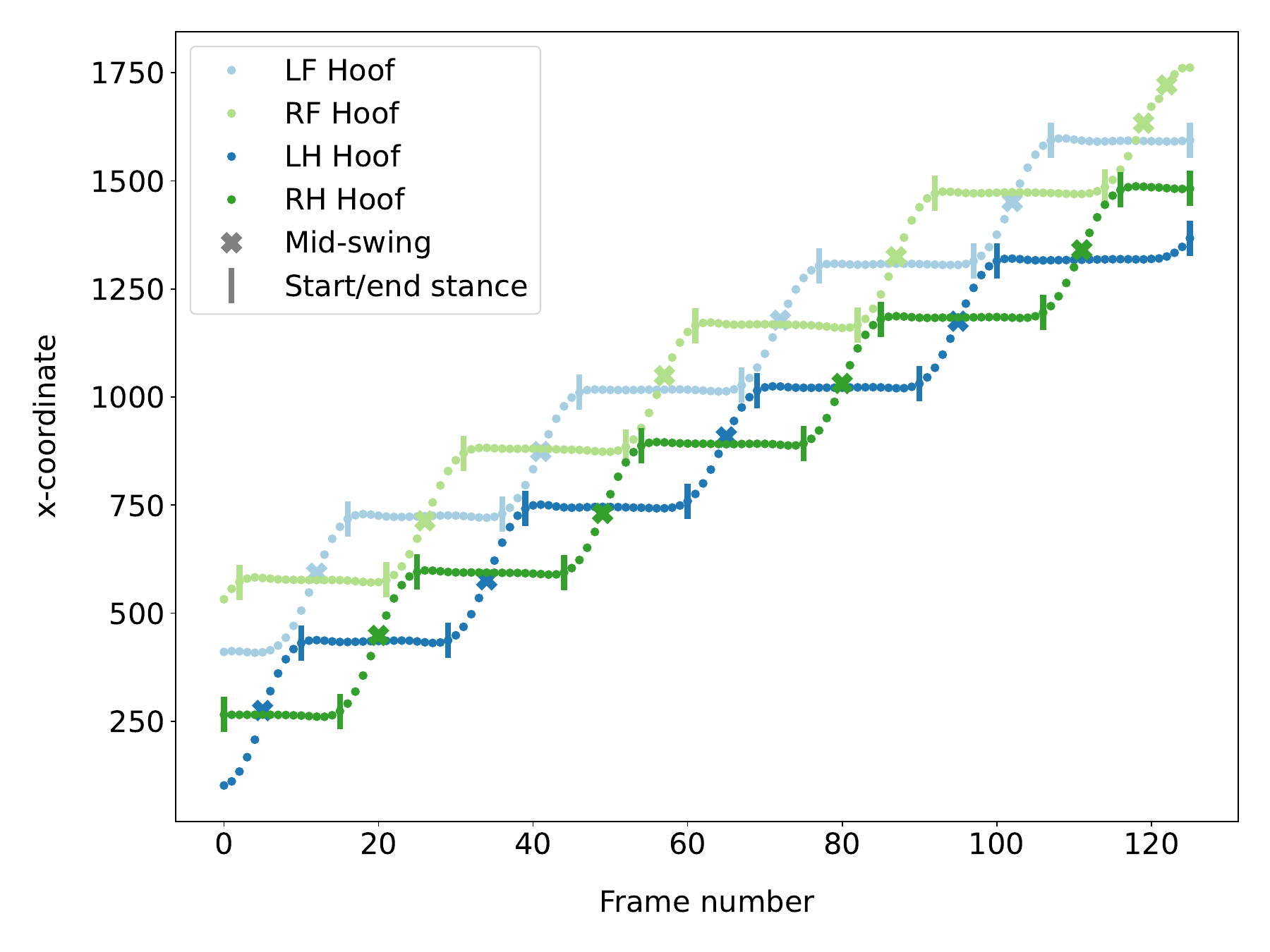}
    \caption{Example of the step detection, using the trajectories of the x-coordinates of the hooves. The vertical lines mark the beginning and end of the stance phase. The crosses mark the peak of the swing phase.}
    \label{fig:steps-example}
\end{figure}

For each leg, the horizontal movement (x-coordinate) of the hoof was used to detect the stance and swing phases. 
The stance phase starts when a hoof lands on the floor and ends when the hoof moves forward again. 
At that moment, the swing phase starts. 
The hoof continues moving forward for the whole duration of the swing phase until it lands and remains still for another stance phase. 
The start and end frames of the stance phases were detected by finding when the x-coordinates of the hoof remained the same, that is, by finding plateaus of at least 10 frames where the absolute difference in x-coordinates between two frames was $\leq$ 10 pixels, to account for small jitters.
We define mid-swing as a frame between the liftoff and landing of the hoof, just before the hoof starts to slow down.
The mid-swing moments were detected by finding the peaks of the acceleration of the x-coordinates. 
The horizontal acceleration of the hoof was computed by taking the second-order derivative of the x-coordinates and then passed through a uniform filter of size 3.
An example of the x-coordinate trajectories is shown in Figure~\ref{fig:steps-example}, with the stance and mid-swing phases identified by the step detection.

\subsubsection{Step correction}
The step detection was automatically controlled and corrected using the following procedure: for any given leg, mid-swings must happen before or after the stance phases, and the mid-swings must happen during the supporting phase of the opposite leg (left-right). 
When the step detection failed to meet these requirements, this indicated that the keypoint predictions were too noisy on that hoof.
Only four videos were found to have problematic step detection.
The frames with problematic steps were then removed from the keypoint trajectories, resulting in trajectories with one or several gaps. 
The trajectories were then trimmed to the part with the most remaining frames.

\subsubsection{Back posture measurement (BPM)}
To estimate the back posture, or curvature of the back, a similar approach as described in~\cite{poursaberi_real-time_2010} was taken. 
A circle was fitted through the three keypoints on the spine. The curvature of a circle can be found by taking the inverse of its radius.
The radius ($r$) of the fitted circle was normalized with the head length ($h$) of the cow (in pixels), as the length of cows can differ. 
The head length was taken as the Euclidean distance between the keypoints on the forehead and the nose.
The BPM was then calculated as follows:

\begin{align}
    \text{BPM} &= \frac{1}{r/h} = \frac{h}{r}
\end{align}
For each leg, the BPM was computed at each mid-swing phase. 
If there were multiple swing phases, the median BPM value was kept for that leg. 
The largest BPM over all four legs was used as the final BPM value.

\subsubsection{Head bobbing amplitude (HBA)}
\begin{figure}[ht]
    \centering
    \begin{subfigure}[b]{0.5\textwidth}
        \centering
        \includegraphics[width=1\textwidth]{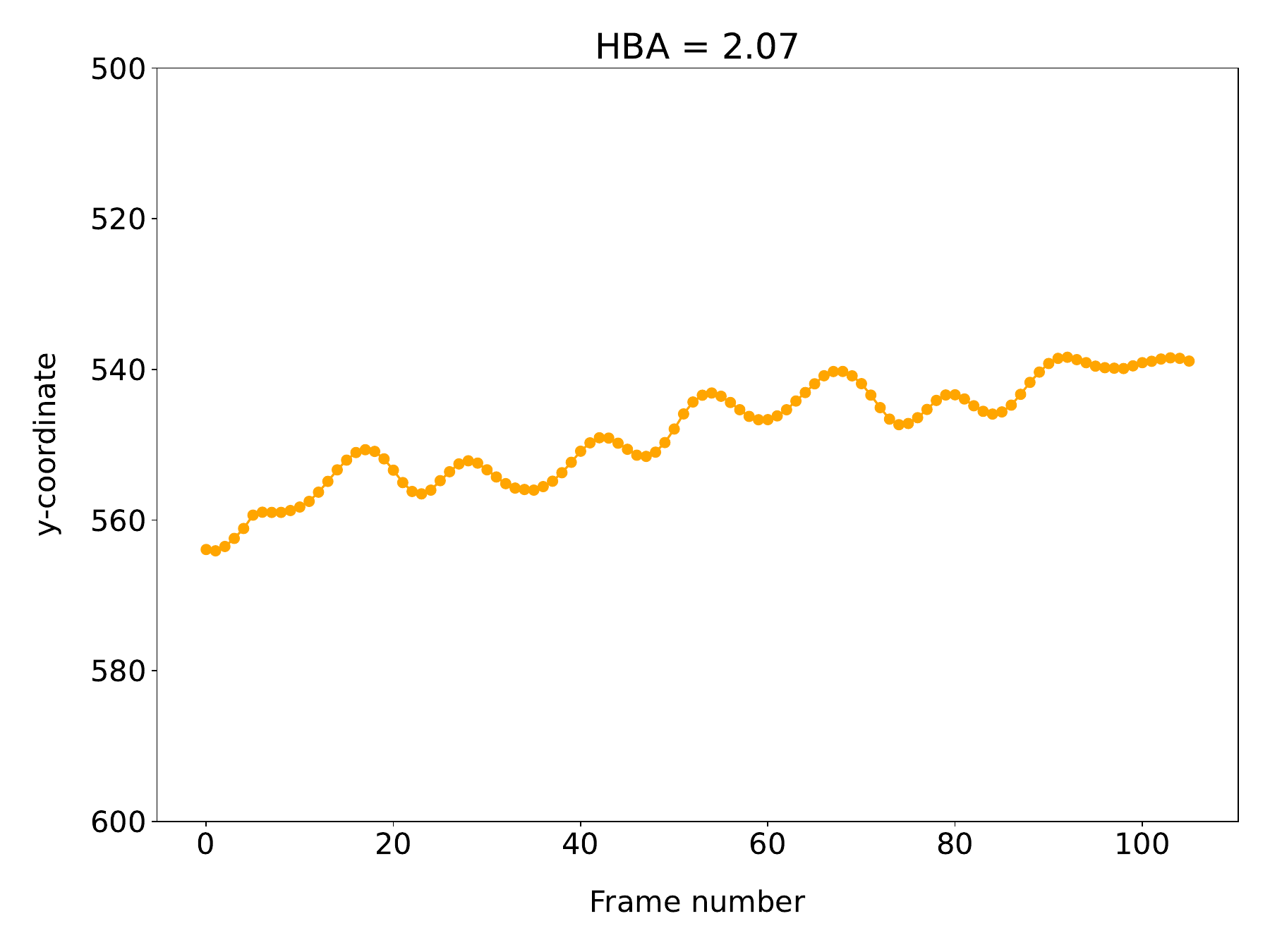}
        \caption{Y-signal of the head keypoint without noticeable head bob.}
    \end{subfigure}%
    \hfill
    \begin{subfigure}[b]{0.5\textwidth}
        \centering
        \includegraphics[width=1\textwidth]{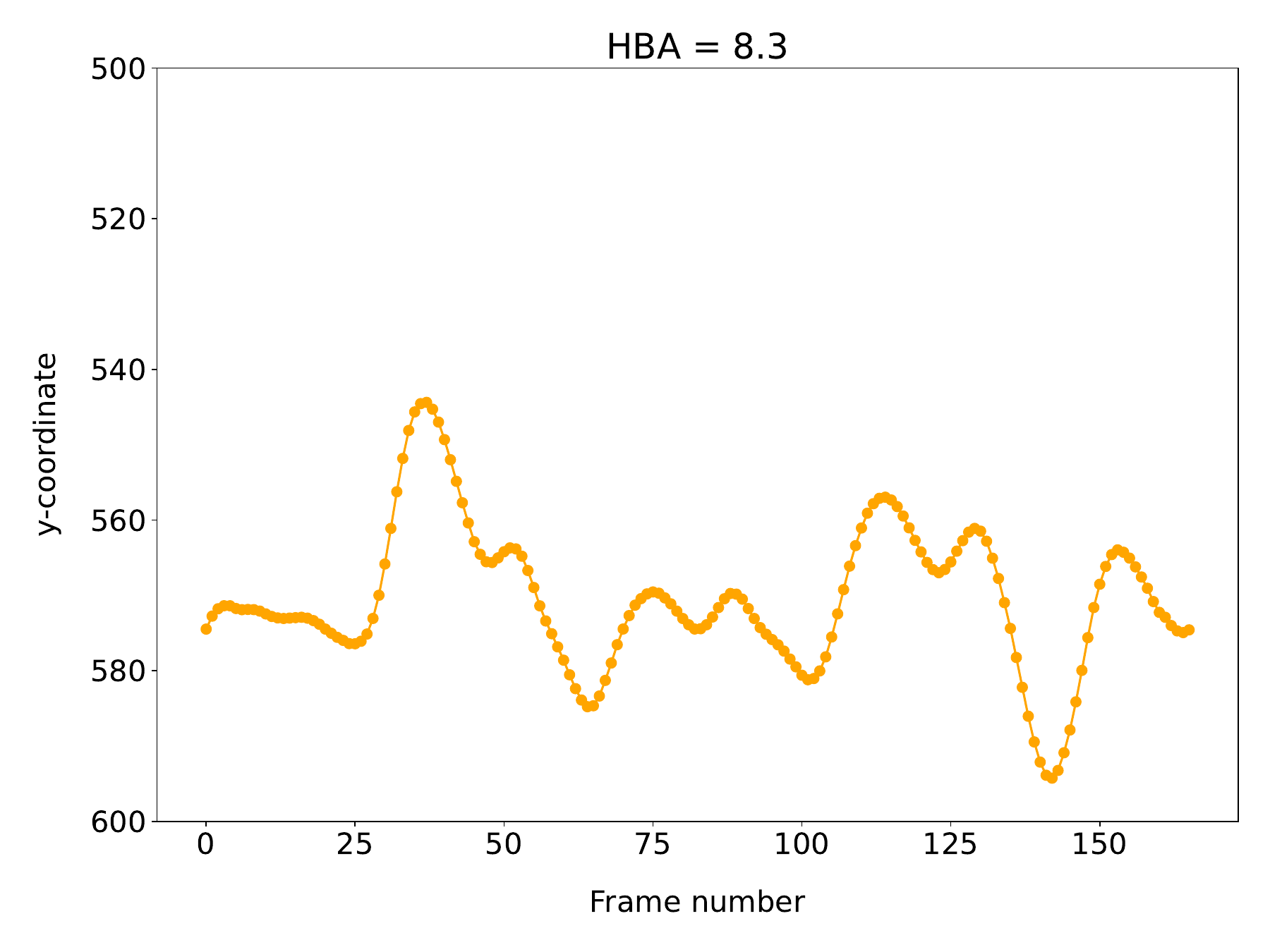}
        \caption{Y-signal of the head keypoint with noticeable head bob.}
    \end{subfigure}%
    \caption{Example of y-signal with and without head bobbing.}
    \label{fig:hba-example}
\end{figure}

Head bobbing is defined as an exaggerated movement of the head when an affected limb lands and lifts from the ground~\citep{schlageter-tello_relation_2015, blackie_associations_2013}. 
Hence, in the presence of head bobbing, the head moves significantly up and down cyclically (at least once per gait cycle). 
Sound subjects are expected to have a more steady head stance. 
Examples of a noticeable head bob and steady head stance are shown in Figure~\ref{fig:hba-example}.
The amplitude of the vertical movement (y-signal) of the forehead keypoint was used as a measure of head bobbing. 
The amplitude of the y-signal was computed with fast Fourier transforms~\citep{cooley1965algorithm} as follows: let $N_v$ be the number of frames in a video, let $N_g$ be the number of frames per gait cycle in a video, $k \in [1, N_v]$ the frequency, $X$ the Fourier transform of the signal, and $A_k$ the amplitude at frequency $k$.  
The value of the HBA was then assigned as the largest amplitude in a gait cycle:

\begin{align}
    A_k &= \frac{|X_k|}{N_v} \\
    \text{HBA} &= \max^{N_g}_{k=0}(A_k) 
\end{align}

\subsubsection{Tracking distance (TRK)}
The tracking distance is defined as the horizontal distance (x-coordinate) between the landing position of the front hoof and the subsequent landing position of the hind hoof of the same side. 
If the hind hoof lands at the same location as the front hoof, it indicates no serious walking problem~\citep{song_automatic_2008}, and the TRK value is equal (or close) to 0. 
The tracking distance was measured on the left (TRK$^L$) and right (TRK$^R$) side of the cow and was normalized to the head length ($h$) as follows:
for any given side (left, right), let $x_f$ and $x_h$ be the x-coordinates of the front and hind hooves, 
Let $s$ be the start frame of a stance phase on the front hoof, and $s+1$ the start frame of the subsequent stance phase on the hind hoof. 
When there was more than one value per side, the median TRK value of that side was returned. 

\begin{equation}
    \text{TRK} = \frac{x_{f_s} - x_{h_{s+1}}}{h}
\end{equation}

\subsubsection{Stride length difference (STL)}
The stride length is defined as the horizontal distance between two successive landings of the same hoof. 
The stride length ($l$) was measured for each hoof between each successive stance phase ($s$) and normalized to the head length ($h$). 
If there was more than one stride length per hoof, the median value was kept.
We measured the difference in stride length between the left and right sides for the hind (STL$^H$) and front (STL$^F$) legs as follows:

\begin{align}
    l_s &= \frac{x_s - x_{s-1}}{h} \\
    \text{STL} &= | l^{\text{right}} - l^{\text{left}} |
\end{align}

\subsubsection{Stance duration difference (STD)}
We define the stance duration as the time (in seconds) between the start ($a$) and end ($b$) of each stance phase.
The time in seconds was derived from the frame rate of the video recording (here, 30 fps).

The stance duration ($t$) was measured per hoof for each stance phase. 
If a leg had more than one stance phase, the median duration was used.
We measured the difference in duration between the left and right sides for the hind (STD$^H$) and front (STD$^F$) legs as follows:

\begin{align}
    t &= b - a \\
    \text{STD} &= | t^{\text{right}} - t^{\text{left}} |
\end{align}

\subsubsection{Swing duration difference (SWD)}
We define the swing duration as the time (in seconds) between the ($a$) and end ($b$) of each swing phase.
The time in seconds was derived from the frame rate of the video recording (here, 30 fps).

The swing duration ($w$) was measured per hoof for each swing phase.
If a leg had more than one swing phase, the median duration was used.
We measured the difference in duration between the left and right sides for the hind (SWD$^H$) and front (SWD$^F$) legs as follows: 

\begin{align}
    w &= b - a \\
    \text{SWD} &= | w^{\text{right}} - w^{\text{left}} |
\end{align}

\begin{figure}[ht]
    \centering
        \includegraphics[height=0.9\textheight]{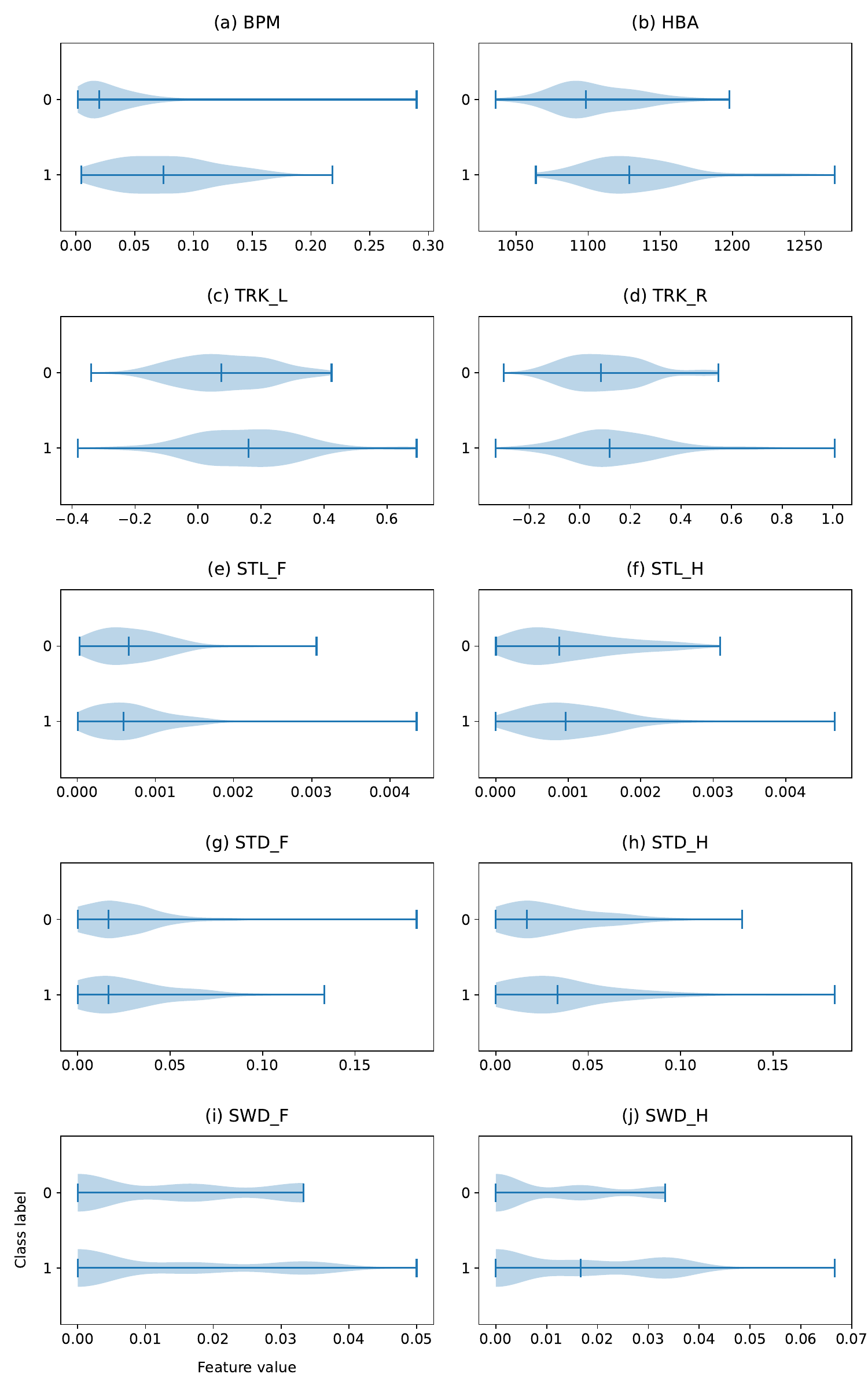}
        
    \caption{Distribution of the features per lameness class, where 0 corresponds to normal, and 1 to lame.}
    \label{fig:feat-distrib}
\end{figure}

A summary of the features extracted is listed in Table~\ref{tab:feat-list}, and Figure~\ref{fig:feat-distrib} presents the distribution of the values of each feature per lameness class.

\begin{table}[ht]
\centering
\caption{List of the features extracted from the keypoint trajectories.}
\label{tab:feat-list}
\begin{tabular}{@{}ll@{}}
\toprule
\textbf{Feature} & \textbf{Description}                                                          \\ \midrule
BPM          & Back posture measurement                                             \\
HBA          & Head bobbing amplitude                                               \\
TRK$^L$        & Tracking distance on the left side                                   \\
TRK$^R$        & Tracking distance on the right side                                  \\
STL$^F$        & Stride length difference between left- and right-front hooves   \\
STL$^H$        & Stride length difference between left- and right-hind hooves     \\
STD$^F$        & Stance duration difference between left- and right-front hooves \\
STD$^H$        & Stance duration difference between left- and right-hind hooves   \\
SWD$^F$        & Swing duration difference between left- and right-front hooves  \\
SWD$^H$        & Swing duration difference between left- and right-hind hooves    \\ \bottomrule
\end{tabular}
\end{table}

\subsection{Lameness classification}

The layout of our machine-learning experiments is described in the next paragraphs.
We first split the data into training and validation sets using cross-validation. 
We then trained and evaluated different binary classifiers to detect lameness using all the extracted features. 
Lastly, we investigated the importance of features on classification performance.

\subsubsection{Data preparation}
Considering the relatively small dataset size (272 videos), the dataset was split into training and validation sets using a 5-fold cross-validation (CV) with stratified grouping.
In order to prevent data leakage, the grouping was performed on the cow IDs to ensure that, in each fold, there was no overlap of cow IDs between the training and the validation set. 
Given this non-overlapping constraint, the stratification creates folds that retain, as much as possible, the same class distribution~\citep{buitinck2013api}. 
To ensure a balanced class distribution during training, we applied the Synthetic Minority Oversampling Technique (SMOTE)~\citep{chawla2002smote} to the minority classes in the training sets. 
SMOTE generates new training samples whose feature values are close to the other samples in the minority class.
Lastly, the features were re-scaled as machine-learning models often require the features to be on a similar scale. 
The range of the features was re-scaled using Robust Scaling~\citep{scikit-learn}, which uses statistics that are robust to outliers for scaling the data. 

\subsubsection{Classification models}
We compared the performance of the following six classifiers: Logistic Regression (LR), Random Forest (RF), Support Vector with a linear kernel (SVL) and with a radial kernel (SVR), Multi-Layer Perceptron (MLP) and Gradient Boosting Machines (GB). 
These classifiers were selected as they showed good performance in previous research on lameness detection~\cite{zhao_automatic_2023, zheng_cows_2023, arazo_segmentation_2022}.
We used a flat cross-validation approach to tune the hyper-parameters and train the models, as it is computationally less expensive than nested cross-validation, and generally results in the selection of an algorithm of similar quality to that selected via nested cross-validation~\cite{wainer2021nested}.
The hyper-parameters of the classifiers were first optimized using a random cross-validated search of 100 iterations over the 5-folds.
The classifiers were then re-trained on the 5-folds with the best set of hyper-parameters.

\subsubsection{Evaluation metrics}
The performance of the classification models was evaluated with the following metrics: accuracy, F1-score, sensitivity, and specificity. The F1-score was macro-averaged; that is, the metric was calculated per class and then averaged. The macro-average is especially useful with imbalanced datasets, as all classes contribute equally to the metric.

\subsubsection{Feature importance}
An additional experiment was run to investigate whether including multiple features could lead to improvements in lameness classification.
The predictive value of a feature was evaluated by measuring the feature importance, that is, how much a feature contributed to a correct classification. 
To measure the feature importance, we selected the permutation importance method~\citep{breiman2001random} as it can be applied to any classifier.
The importance of features was evaluated on the best-performing classifier among the 6 classifiers that were trained with all the features.
The permutation importance method was performed as follows:
For each cross-validation fold, the model was fitted on the training dataset and evaluated on the F1-score on the validation set.
Then, a feature column from the validation set was randomly shuffled, and the model was evaluated again.
The importance score was then the difference between the F1-score on the non-shuffled and the shuffled validation data.
The permutations were repeated 100 times for each feature.
The features were then ranked in the order of their mean importance score.
To estimate whether including multiple features could lead to improvements in the lameness classification, the classifier was then retrained with the most important feature, the two most important features, and so on, gradually adding one feature in the order of their importance.

\section{Results}

\subsection{Pose estimation}
The test results of T-LEAP are presented in Table~\ref{tab:res-tleap}. On average, there were 99.6\% of correctly detected keypoints (PCKh@0.2). In other words, the Euclidean distance between the predicted keypoint and its ground truth was smaller than 20\% of the head length in 99.6\% of the cases. This is in line with the results presented in the original paper~\citep{russello_t-leap_2021}, where they achieved a 99.0\% detection rate on the same model with 17 keypoints. 
The keypoint correction and filtering were run on all 272 videos, and the MAD filter (of window size 3) identified 0.21\% of outlier keypoints, whose coordinates were then corrected to the median value of the temporal window.
Because of the lack of keypoint annotations on all videos, the keypoint correction could only be assessed qualitatively. 
The trajectories of the keypoints before and after the filtering were plotted for each video and controlled visually. 
The quality of the filtered trajectories was deemed balanced, in that most of the outliers could be corrected and the trajectories appeared smooth, without over-correction or flattening.
The outliers that could not be corrected sufficiently led to a wrong step detection. These steps were then discarded from trajectories, as detailed in section~\ref{subsec:feat-extr}.

\begin{table}[h]
    \centering
    \caption{Percentage of Correct Keypoints (PCKh@0.2) of T-LEAP on the test set. 
    The keypoints are named as follows: 1: Left-hind hoof, 2: Right-hind hoof, 3: Left-front hoof, 4: Right-front hoof 5: Nose, 6: Forehead, 7: Withers, 8: Sacrum, 9: Caudal thoracic vertebrae.}
    \label{tab:res-tleap}
    \begin{tabular}{@{}lcccccccccc@{}}
    \toprule
    \textbf{Keypoint} & \textbf{1} & \textbf{2} & \textbf{3} & \textbf{4} & \textbf{5} & \textbf{6} & \textbf{7} & \textbf{8} & \textbf{9} & \textbf{Mean} \\ \midrule
    PCKh@0.2 & 98.45 & 1 & 99.48 & 98.45 & 100 & 100 & 100 & 100 & 100 & 99.60 \\ \bottomrule
    \end{tabular}
\end{table}

\subsection{Lameness detection}
The results of the different binary classifiers are listed in Table~\ref{tab:res-all}.
The SVM with radial kernel, Random Forests, and Gradient Boosting classifiers performed best, with an accuracy above 79\%. 
SVM-R had a higher specificity, while the Random Forests and Gradient Boosting had a higher sensitivity.
The logistic regression, the SVM with linear kernel, and the Multi-Layer Perceptron performed slightly worse.

\begin{table}[h]
    \centering
    \caption{Results of the binary classifiers using all the features. Values are expressed in \%. The best results are highlighted in bold.}
    \label{tab:res-all}
    
    \begin{tabular}{@{}lcccc@{}}
    \toprule
    \textbf{\small{Model}}         & \textbf{\small{Accuracy}}     & \textbf{\small{F1-score}}     & \textbf{\small{Sensitivity}}    & \textbf{\small{Specificity}}       \\ \midrule
    Logistic Regression    & 78.49                & 77.26                & 77.33                  & 77.90                \\
    SVM linear kernel   & 77.25                & 76.31                & 75.39                  & 77.90                \\
    SVM radial kernel   & \textbf{80.07}       & \textbf{78.70}       & 76.78                  &  \textbf{81.15} \\
    Random Forests         & 79.66                & 78.44                & 83.68                  & 74.64          \\
    Gradient Boosting      & 79.12                & 77.79                & \textbf{84.60}         & 72.05          \\
    Multi-Layer Perceptron & 78.97                & 77.60                & 80.74                  & 74.59                \\ \bottomrule
    \end{tabular}
\end{table}

\subsection{Feature importance}
\begin{figure}[h]
	\centering
	\includegraphics[width=0.8\textwidth]{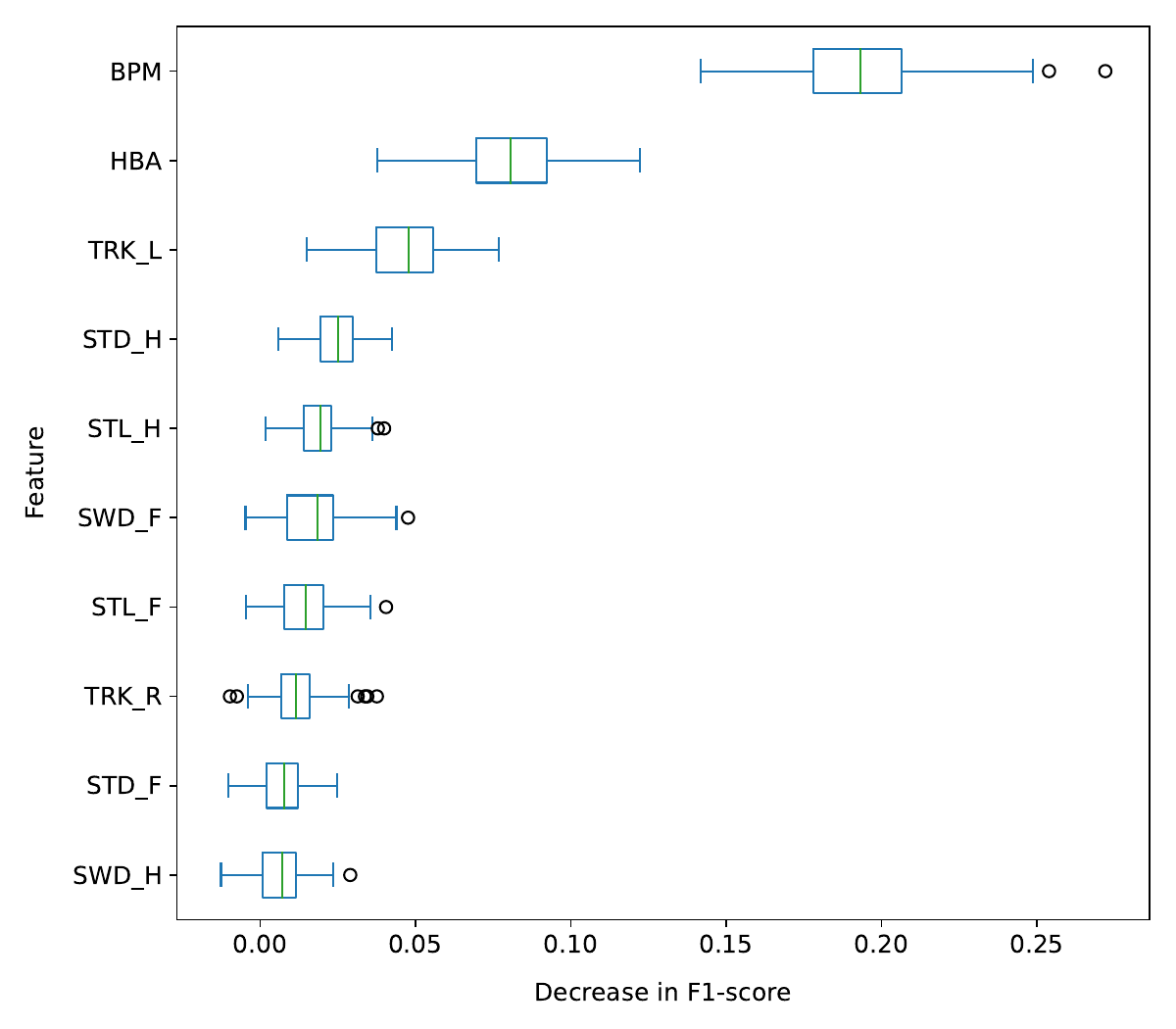}
	\caption{Results of the feature importance over 100 random permutations.}
	\label{fig:feat-importance}
\end{figure}

A plot with the scores returned by the permutation importance is shown in Figure~\ref{fig:feat-importance}.
For each feature, the score indicates how much a random permutation of the feature values impacted the prediction scores, averaged over 100 permutations.
The Back Posture Measurement (BPM) had the highest permutation score, followed by the Head Bobbing Amplitude (HBA) and Left Tracking Distance (TRK\_L). 
The remaining features showed less importance.

Using the permutation importance results, the SVM classifier with radial kernel (SVM-R) was then retrained by gradually adding one feature, in the order of their importance. 
The classification results of the classifier using these different combinations of features are presented in Table~\ref{tab:res-ffs}.
In terms of accuracy and F1-score, using two or more features improves the classification results compared to only using BPM.
The best classification scores are reached by using combinations of 3 and 6 features.

\begin{table}[ht]
    \centering
    \caption{Results (in \%) of the SVM-R classifier after gradually adding one feature per order of their importance score.}
    \label{tab:res-ffs}
    \begin{tabular}{@{}lcccc@{}}
    \toprule
    \textbf{\small{SVM-R Features}}             & \textbf{\small{Accuracy}} & \textbf{\small{F1-score}} & \textbf{\small{Sensitivity}}  & \textbf{\small{Specificity}}   \\ \midrule
        \scriptsize{BPM}                             & 76.66            &  74.81           &  63.26               &  86.69  \\
        \scriptsize{BPM, HBA}                        & 79.31            &  77.50           &  77.42               & 77.32 \\
        \scriptsize{BPM, HBA, TRK}                   & 79.87            & 78.22            & 76.35                & 80.14     \\ 
        \scriptsize{BPM, HBA, TRK, STD}              & 79.47            & 77.87            & 77.09                &  78.89    \\ 
        \scriptsize{BPM, HBA, TRK, STD, STL}         &  79.18           & 78.03            &  78.31               &    79.17  \\ 
        \scriptsize{BPM, HBA, TRK, STD, STL, SWD}    &  80.07           &  78.70           &  76.78               &  81.15      \\ 
    \bottomrule
    \end{tabular}
\end{table}

\section{Discussion}

\subsection{Video processing}
The video processing consisted of the following steps: using Faster-R-CNN to detect and isolate the cows from the video frames, using T-LEAP to extract time-series of keypoint locations, and using the MAD and Savitzky–Golay filters to reduce noise from the keypoint predictions.
For our set of videos, the pre-trained Faster-R-CNN worked out of the box and detected the location of the cows in each video frame.
The performance of T-LEAP was on par with the results described in the original paper~\citep{russello_t-leap_2021}, and it would require little effort to be transferred to videos recorded in new farms, as~\cite{taghavi2023cow} showed that little new training data was needed to fine-tune the T-LEAP model.
However, some keypoint mis-detections needed to be corrected.
The parameters for the MAD outlier filter and the smoothing Savitzky–Golay filter had to be tuned manually until a good trade-off between under- and overcorrection was found. 
With no or insufficient correction of the keypoint trajectories, the features could give erroneous values.
While with over-correction, one would run the risk of removing the true signal of keypoint trajectories, and the extracted features wouldn't be discriminatory. For instance, if the signal of the forehead would be too flattened, the head bobbing would be systematically missed.

The videos were selected such that there was only one cow at a time in the field of view. 
This constraint makes the gait analysis more reliable in two ways. 
First, having a single cow in the field of view ensures that the cows don't occlude each other's body parts, making the pose estimation more reliable.
Second, a single cow in the field of view ensures enough space between the cows such that they can walk at their own pace and display a voluntary gait. 
In practice, this constraint could be implemented by skipping the videos where the Faster-R-CNN (or any other object detector) detects more than one cow, or as done in~\cite{barney2023deep}, by implementing a tracking algorithm that follows each cow through the video.

Another constraint for selecting the video was that the cows had to walk from left to right.
However, our proposed method would still work if the cows were walking in the opposite direction. 
We can determine the walking direction by examining the horizontal movement of the keypoints along the x-axis. The values of the keypoints x-coordinates increase over time when the cows walk to the right and decrease when they walk to the left. 
Therefore, if the cows were to walk from right to left, their keypoint trajectories could simply be mirrored on the x-axis before extracting the features.

\subsection{Locomotion scoring}

A classifier learns to classify samples from a set of labeled examples, also known as ground-truth or golden-standard. 
Because a classifier can only be as accurate as its golden-standard~\citep{schlageter-tello_relation_2015}, a reliable locomotion scale is necessary.
Here, the initial inter- and intra-observer reliability was under par. 
It is worth noting that the reliability is usually lower in homogeneous data because the probability of agreement by chance is higher when scores are not equally distributed~\citep{schlageter-tello_effect_2014}.   
It is unlikely that scoring from live observations instead of from videos would have improved the scores, as~\cite{schlageter2015comparison} showed no difference in the reliability of inexperienced observers between live and video scoring and showed improved reliability of experienced observers when scoring from video.
The quality of the ground-truth could perhaps have been further improved by organizing additional locomotion scoring sessions or by having shorter scoring sessions over multiple days.
However, given that the availability of the observers was limited and that a perfect golden standard was not necessary nor likely achievable, we took other steps to address the problem of low reliability and agreement. 
Firstly, because we had multiple observers, we could discard the votes from the least reliable observers. 
Secondly, we addressed the problem of class (score) imbalance by merging the levels of the scale to a binary score: normal and lame.
Although these steps improved the quality of our golden standard, some biases might have remained, which could limit the classifiers' accuracy.

\subsection{Lameness detection}

The lameness detection was performed as a binary classification task (\textit{normal} vs. \textit{lame}) and therefore focused on lameness detection rather than fine-grained gait scoring.
Fine-grained locomotion scoring is left for future research as it would require collecting more video footage with sufficient examples of gait scores of 3 and above.

The performance of the linear classifiers (i.e., logistic regression and SVM with linear kernel) was lower than that of the non-linear classifiers. 
This implies that when combining all the features, the decision boundary between the \textit{normal} and \textit{lame} classes is non-linear. 
The Multi-Layer Perceptron didn't perform as well as the other non-linear classifiers, most likely because of the relatively small dataset.
The performance of the three best classifiers SVM-R, RF, and BG, aligns with the conclusions of~\cite{wainer2016comparison} and~\cite{wainer2021nested}: they found these three binary classifiers to perform the best on 115 open-source datasets tackling a variety of real-world problems in medicine and biology (but not related to lameness detection).
Although, on this dataset, the SVM classifier with radial kernel achieved the best performance in terms of accuracy and F1-score, it might not be the case for other datasets. This is a well-known machine-learning challenge, also known as the “no-free-lunch” theorem, that suggests that no algorithm can outperform all others for all problems~\citep{wolpert1996lack}.
Our recommendation would then be to try several classifiers, and the SVMs with radial kernel, random forests, and gradient boosting classifiers provide a good starting point.

\subsection{Feature importance}

Multiple studies investigated the relationship between individual locomotion traits and locomotion scores~\citep{flower_effect_2006, borderas2008effect, chapinal2009using, schlageter-tello_relation_2015}.
They found that, when scored individually, the traits arched back, asymmetric gait, head bobbing, reluctance to bear weight and tracking-up were highly correlated with the locomotion score.
The features selected in this study were designed to measure the same traits. The arched back was measured by the Back Curvature Measurement (HBA),  the asymmetric gait by the Stride Length (STL) difference between left and right limbs, the head bobbing by the Head Bobbing Amplitude (HBA), the reluctance to bear weight by the Stance Duration (STD) and Swing Duration (SWD), and the tracking up was measured by the Tracking distance (TRK).

The BPM, HBA, and TRK features returned the highest scores in the permutation importance test. 
BPM and HBA displayed a clear demarcation between the normal and lame classes in Figure~\ref{fig:feat-distrib}.
As reported by~\cite{flower_effect_2006},~\cite{borderas2008effect} and~\cite{chapinal2009using}, it suggests that the back posture, head bobbing, and tracking-up are, for human observers, easier to recognize than an asymmetric gait (e.g. stride length).
The tracking distance on the left side (TRK-L) had a higher importance than the one on the right side (TRK-R). 
This could indicate that, in our dataset, there were more cows tracking-up on the left than on the right side.

Both for the Stance Duration (STD) and the Swing Duration (SWD) on the hind legs (Fig.~\ref{fig:feat-distrib}), one can see a clear difference in the duration of the stance/swing phases between the classes, whereas classes differences are less obvious on the front legs. This could be explained by the fact that lameness happens more often on the hind legs~\citep{poursaberi_real-time_2010, flower_effect_2006}.
Including SWD as a feature increased the classification performance, even though SWD had the lowest importance score. 
In contrast, STD had a larger importance score than SWD, but adding the STD feature to the input of the classifier led to a small decrease in accuracy and F1-score.
This could indicate multi-collinearity with other features.

The STL features had the second lowest importance score and the class separation was harder to distinguish in Figure~\ref{fig:feat-distrib}.
Interestingly, the F1-score, sensitivity, and specificity were higher when the STL features were included.
This suggests that the stride length can be informative when used in combination with other features.
It is worth noting that if the cows have bilateral lameness, i.e., are lame on left and right limbs, then the stride length would show little to no difference~\citep{blackie_associations_2013}.

Overall, combining multiple locomotion traits led to a better classification performance than using a single trait. 
Using a combination of 3 and 6 traits led to the best accuracy and F1-scores on the SVM classifier with a radial kernel. 
Even though additional traits could be extracted from the keypoint trajectories, it is unknown whether they would lead to significant improvements in the gait classification. 
Our recommendation would be to include at least the following locomotion traits in an automatic lameness detection system: back posture, head bobbing, and tracking distance, as they demonstrated good overall classification metrics, and these features have been shown to be highly correlated with the locomotion scores~\citep{flower_effect_2006, borderas2008effect, chapinal2009using}.

\subsection{Comparison with related work}

Directly comparing the performance of our lameness classifiers against related work is not straightforward, because even though the task at hand (i.e., detecting lameness from videos) is the same, there is a large variation in the material, methods, and evaluations used in papers that address it. 
Furthermore, a comprehensive literature review is out of the scope of this paper, and we refer the reader to~\cite{nejati2023technology} for an overview of past and current advances in bovine gait analysis. 
We will here compare our results and contrast our findings with previous work that we deem directly related to ours.

The Back Posture Measurement (BPM) was first introduced by~\cite{poursaberi_real-time_2010} and curvature of the back has since then been used in numerous studies~\citep{poursaberi_real-time_2010, viazzi_analysis_2012, van_hertem_automatic_2014, viazzi_comparison_2014, van_hertem_implementation_2018, jiang_dairy_2022, zheng_cows_2023, zhao_automatic_2023, barney2023deep}. 
The BPM is commonly measured during the supporting phase of the hind hooves, and not during the supporting phase of the front hooves because lameness is more common on the hind hooves than on the front ones.
However, this practice could lead to the algorithm systematically missing front lameness cases.
To prevent this, we computed the BPM based on the supporting phase of the four legs.
When using BPM as a single locomotion trait, the accuracy of lameness classification ranged from 76\%~\citep{viazzi_analysis_2012} to 96\%~\citep{jiang_dairy_2022}.
When only including the BPM feature in our SVM-R classifier, we reached an accuracy of 76.6\%, which is in line with the literature.

The work presented in~\cite{zhao_automatic_2023} is perhaps the most closely related to this study.
In~\cite{zhao_automatic_2023}, the authors used a combination of traditional and deep-learning-based computer vision to develop a lameness detection system.
They used DeepLabCut~\citep{mathis_deeplabcut_2018}, a deep-learning model trained to track the location of the hoofs and the head in videos of walking cows without physical markers. 
The pose estimation model achieved a percentage of correct keypoints (PCK) of 92\% and the videos where the keypoint predictions were too erroneous were manually discarded.
The outline of the cow's spine was obtained using a background subtraction method at the pixel level. 
However, this method may not be effective in handling varying backgrounds and lighting conditions. 
On the other hand, we used pose estimation to track all the keypoints on the cow's body, including those on the spine. 
As a result, our proposed method is more robust in dealing with such changes. 
Additionally, we automatically corrected erroneous keypoint trajectories with our keypoint-correction and step-correction approaches, which removed the need for manual validation.

In total, they used 212 videos of walking cows, where cows with a score of 1 or 2 were classified as \textit{normal}, and a score of 3 or 4 as \textit{lame}.
The back curvature was computed from the outline of the spine, and the keypoints on the hooves and on the neck were used to extract the following features: head bobbing, stride length asymmetry, tracking up, landing speed, supporting phase asymmetry, and moving speed.
The back curvature and head bobbing were computed when the hind hooves were in contact with the floor and, therefore, didn't account for lameness on the front limbs.
In contrast, we computed the back curvature over all four legs, and the head bobbing amplitude was extracted from the entire trajectory. 
The feature selection was performed as follows: a Chi-square test was run on the whole dataset. 
The test revealed that back posture measurement and head bobbing were the most important features.
In contrast, we found that adding tracking-up to the other two features led to better results on our dataset.
This could mean that, in their dataset, lame subjects were not tracking up. 
Another explanation could be that as the number of traits increases, the complexity of the data increases, and a non-linear classifier, such as SVM-R, would be needed.
Several classifiers were trained with the back curvature and head bobbing, and the logistic regression classifier returned the best results, with a classification accuracy of 87.3\%.
In comparison, our accuracy of 80.07\% may seem modest. 
However, it's important to note that there are several differences in the data and experimental setup that could have impacted the results. 
The datasets and methods used in the studies differed, making a direct comparison challenging. 
In our study, most of the locomotion scores were on the lower end of the scale, and the small number of severely lame cows may have made it harder for the classifier to distinguish between lame and non-lame cows. Additionally, the quality of the ground truth data might have affected the performance of our classifiers. 
We also followed best machine-learning practices by conducting feature selection only on the training sets and by separating individual cows in the training and validation sets to prevent data leakage. 
Failing to do so can artificially inflate performance results and lead to overly optimistic conclusions~\citep{wang2019validation}.
Overall, we would expect our method to yield similar results if it were run on their dataset, with the advantage of a fully automatic pipeline that doesn't require manual validation of the keypoint trajectories, and a pose estimation method robust to light conditions and occlusions.

In~\cite{barney2023deep}, a fully automated multi-cow lameness detection system was developed. 
They used a Mask-R-CNN, a deep-learning model, to simultaneously perform object-detection of the cows, and pose estimation of 7 keypoints on the back neck and head. 
In total, they used 250 videos of 10 different cows.
The keypoints were used to extract the back curvature and head position locomotion traits.
Each locomotion trait was extracted per video frame and aggregated per video into statistical features such as the mean, median, standard deviation, min, and max values.
They trained the CatBoost gradient boosting classifier and achieved a 98\% accuracy on binary lameness detection and 94\% accuracy on a 4-point scale lameness scoring.
In our work, although we included four more locomotion traits, we only aggregated the values into the median value of the video.
In light of their classifiers' excellent performance, a promising direction for extending our work would be to extract more statistical features from the locomotion traits, such as mean, standard deviation, and min and max values, to improve our classification performance further.

\section{Conclusion}
In this paper, we developed a fully automated lameness detection system.
Using the T-LEAP pose estimation model, the motion of nine keypoints was extracted from videos of walking cows.
The trajectories of the keypoints were then used to compute six locomotion traits, namely back posture measurement, head bobbing, tracking distance, stride length, stance duration, and swing duration. 
We found that the three most important traits were back posture measurement, head bobbing, and tracking distance and that including multiple locomotion traits led to a better classification than with a single locomotion trait.
For the ground truth, we showed that a thoughtful merging of the scores of the observers could improve intra-observer reliability and agreement.
Future work should evaluate the system in a less constrained environment, for instance, with multiple cows in the field of view.
Another area for future research could focus on leveraging the temporal essence of the videos, by for instance, including more statistical features per locomotion traits.%

\section*{Acknowledgements}
This publication is part of the project Deep Learning for Human and Animal Health (with project number EDL P16-25-P5) of the research program Efficient Deep Learning (\url{https://efficientdeeplearning.nl}) which is (partly) financed by the Dutch Research Council (NWO).

\section*{Declaration of interests}
The authors declare that they have no known competing financial interests or personal relationships that could have appeared to influence the work reported in this paper.

\clearpage
\bibliography{gait-analysis-paper.bib}

\end{document}